\documentclass[twoside]{article}
\usepackage{times}
\usepackage{epsfig}
\usepackage{graphicx}
\usepackage{amsmath}
\usepackage{amssymb}

\usepackage{cuted}
\usepackage{IEEEtrantools}
\usepackage{amssymb,amsmath}%
\usepackage{amsthm}
\usepackage{mathtools}
\usepackage{wrapfig}
\usepackage{scrextend}
\usepackage{times}
\usepackage{xcolor}
\usepackage{bm}
\usepackage[commentsnumbered]{algorithm2e}
\usepackage{amsthm}
\usepackage{url}

\usepackage{subcaption}
\usepackage{booktabs}  
\usepackage{diagbox}

\usepackage{amsmath,amsfonts,bm}

\def\eqref#1{equation~\ref{#1}}

\def\1{\bm{1}}

\DeclareMathAlphabet{\mathsfit}{\encodingdefault}{\sfdefault}{m}{sl}
\SetMathAlphabet{\mathsfit}{bold}{\encodingdefault}{\sfdefault}{bx}{n}

\DeclareMathOperator*{\argmax}{arg\,max}

\newcommand{\bcdot}{\boldsymbol{\cdot}}
\newcommand{\eqdef}{\mathrel{\mathop:}=}

\newcommand{\exptt}[2]{\mathbb{E}_{#1}\left[#2\right]}
\newcommand{\matr}[1]{\mathbf{#1}}
\renewcommand{\vec}{\boldsymbol}

\usepackage[accepted]{aistats2022_arxiv}
\usepackage[hidelinks]{hyperref}

\usepackage[round]{natbib}

\begin{document}

\twocolumn[

\aistatstitle{Learning to Attack with Fewer Pixels: A Probabilistic Post-hoc Framework for Refining Arbitrary Dense Adversarial Attacks}

\aistatsauthor{He Zhao\IEEEauthorrefmark{1}, Thanh Nguyen\IEEEauthorrefmark{1}, Trung Le\IEEEauthorrefmark{1} 
\AND Paul Montague\IEEEauthorrefmark{2}, Olivier De Vel\IEEEauthorrefmark{2}, Tamas Abraham\IEEEauthorrefmark{2}
\AND Dinh Phung\IEEEauthorrefmark{1}}

\aistatsaddress{\IEEEauthorrefmark{1}Monash University, Australia \AND \IEEEauthorrefmark{2}Defence Science and Technology Group, Department of Defence, Australia } ]

\begin{abstract}
Deep neural network image classifiers are reported to be susceptible to adversarial evasion attacks, which use carefully crafted images created to mislead a classifier.
Many adversarial attacks belong to the category of dense attacks, which generate adversarial examples by perturbing all the pixels of a natural image.
To generate sparse perturbations, sparse attacks have been recently developed, which are usually independent attacks derived by modifying a dense attack's algorithm with sparsity regularisations, resulting in reduced attack efficiency. In this paper, we aim to tackle this task from a different perspective. We select the most effective perturbations from the ones generated from a dense attack, based on the fact we find that a considerable amount of the perturbations on an image generated by dense attacks may contribute little to attacking a classifier. Accordingly, we propose a probabilistic post-hoc framework that refines given dense attacks by significantly reducing the number of perturbed pixels but keeping their attack power, trained with mutual information maximisation. Given an arbitrary dense attack, the proposed model enjoys appealing compatibility for making its adversarial images more realistic and less detectable with fewer perturbations. Moreover, our framework performs adversarial attacks much faster than existing sparse attacks.
\end{abstract}

\section{INTRODUCTION}
Recently, Deep Neural Networks (DNNs) have enjoyed great success in many application areas such as computer vision and natural language processing. Nevertheless, DNNs have been demonstrated to be vulnerable to adversarial attacks with samples crafted deliberately to fool DNN classifiers~\citep{goodfellow2014explaining,nguyen2015deep,kurakin2016adversarial}. For example, in image classification, an adv example may be perceived as a legitimate data sample in a ground-truth class but misleads a DNN classifier to predict it into a maliciously-chosen target class or any class different from the ground truth. Such adversarial attacks result in severe fragility of machine/deep learning systems, which lead to significant research effort on defending them~\citep{zhang2019theoretically,bui2020improving,bui2021improving,bui2022a,le2022_global_local,nguyen-duc2022particle}.

The most common way to generate adv examples is by adding small perturbations/noise to the pixels of a real image, where two important factors need to be taken into account: perturbation magnitude and location~\citep{sapfECCV2020}.
As adv examples are required to challenge classifiers but be less perceptible to human eyes, the two factors need to be controlled judiciously.
There have been two popular categories of adversarial attacks: \textit{dense attacks} and \textit{sparse attacks}.
For a dense attack, it tries to apply perturbations with small magnitude and usually cares less about the perturbation locations. Actually, many widely-used dense attacks perturb {\em all} the pixels of an image, according to the gradient of the classifier loss for each pixel, e.g., in~\cite{goodfellow2014explaining,moosavi2016deepfool,carlini2017towards,xiao2018generating,athalye2018obfuscated,croce2020reliable}.
Alternatively, a sparse attack seeks fewer locations (pixels) to perturb.
It is shown that by finding appropriate locations, one can change a classifier's prediction of an image by perturbing only a few of its pixels~\citep{papernot2016limitations,narodytska2016simple,schott2018towards,su2019one,modas2019sparsefool,croce2019sparse,sapfECCV2020}. 

Although sparse attacks can attack with fewer perturbations, their perturbations are usually of a larger magnitude than dense attacks, which may result in more perceptible adversarial examples. One may consider a compromise solution that allows a sparse attack to perturb more (but not all) pixels whilst simultaneously reducing the magnitude of each pixel. 
Unfortunately, this can be infeasible as a sparse attack requires an individual search process of the locations to perturb for each image, which usually leads to a much slower attack speed than dense attacks. For many sparse attacks, increasing the number of perturbations will significantly reduce their attack speed~\citep{papernot2016limitations,modas2019sparsefool}, making them less useful in practice.

\begin{figure}
        \centering
        \resizebox{0.47\textwidth}{!}{
         \begin{subfigure}[b]{0.16\linewidth}
                 \centering
                 \caption{}
                 \includegraphics[width=0.99\textwidth]{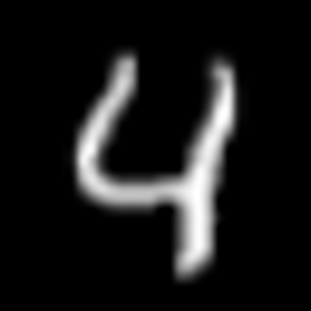}
         \end{subfigure}

         \begin{subfigure}[b]{0.16\linewidth}
                 \centering
                 \caption{}
                 \includegraphics[width=0.99\textwidth]{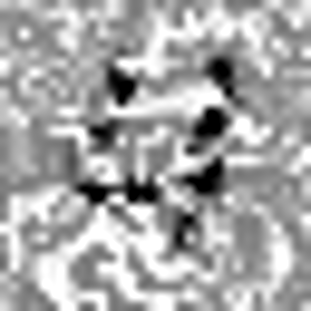}
         \end{subfigure}%
          \begin{subfigure}[b]{0.16\linewidth}
                 \centering
                 \caption{}
                 \includegraphics[width=0.99\textwidth]{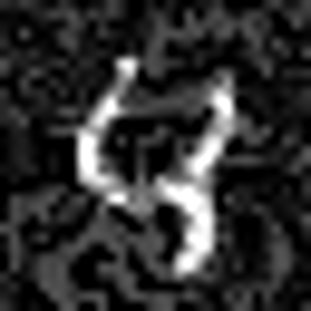}
         \end{subfigure}
         \begin{subfigure}[b]{0.16\linewidth}
                 \centering
                \caption{}
                 \includegraphics[width=0.99\textwidth]{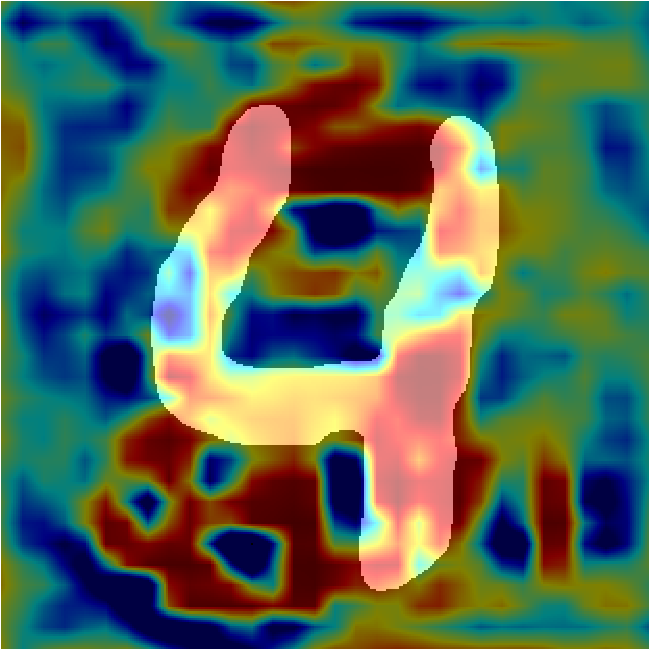}
         \end{subfigure}
          \begin{subfigure}[b]{0.16\linewidth}
                 \centering
                \caption{}
                 \includegraphics[width=0.99\textwidth]{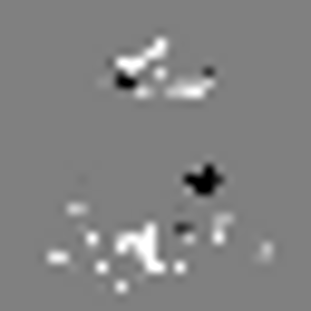}
         \end{subfigure}
       \begin{subfigure}[b]{0.16\linewidth}
                 \centering
                \caption{}
                 \includegraphics[width=0.99\textwidth]{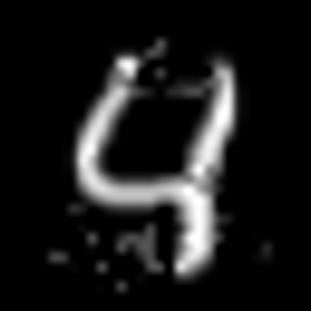}
         \end{subfigure}
         }
         \\
         \vspace{0.2cm}
         \resizebox{0.47\textwidth}{!}{
                  \begin{subfigure}[b]{0.16\linewidth}
                 \centering
                 \caption{}
                 \includegraphics[width=0.99\textwidth]{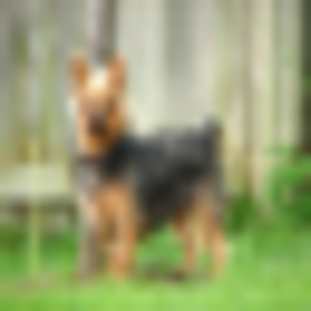}
         \end{subfigure}

         \begin{subfigure}[b]{0.16\linewidth}
                 \centering
                 \caption{}
                 \includegraphics[width=0.99\textwidth]{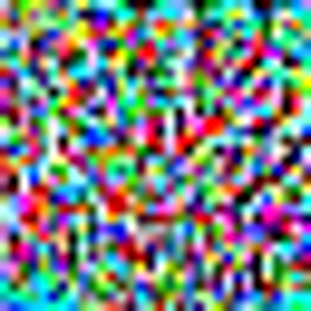}
         \end{subfigure}%
          \begin{subfigure}[b]{0.16\linewidth}
                 \centering
                 \caption{}
                 \includegraphics[width=0.99\textwidth]{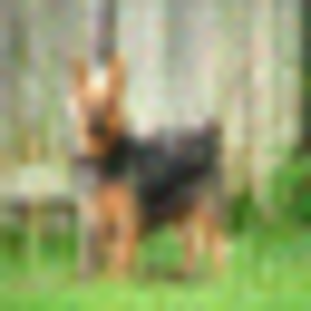}
         \end{subfigure}
         \begin{subfigure}[b]{0.16\linewidth}
                 \centering
                \caption{}
                 \includegraphics[width=0.99\textwidth]{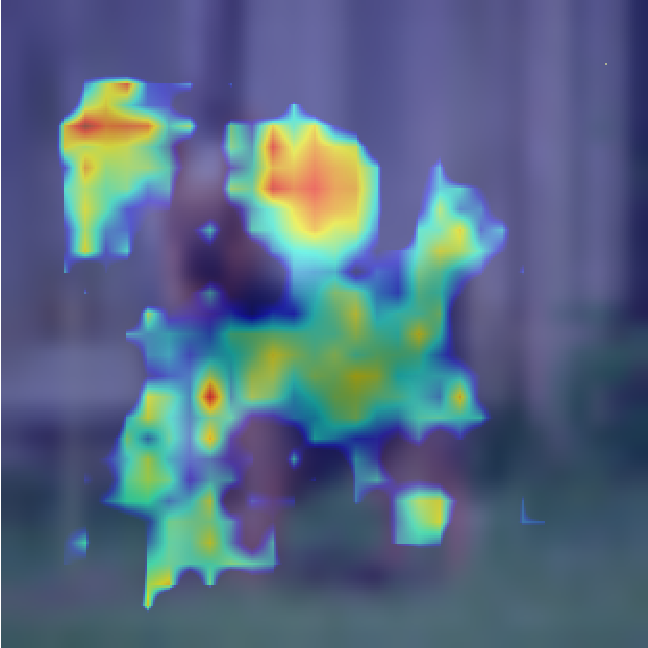}
         \end{subfigure}
          \begin{subfigure}[b]{0.16\linewidth}
                 \centering
                \caption{}
                 \includegraphics[width=0.99\textwidth]{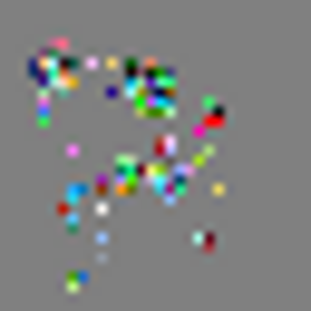}
         \end{subfigure}
       \begin{subfigure}[b]{0.16\linewidth}
                 \centering
                \caption{}
                 \includegraphics[width=0.99\textwidth]{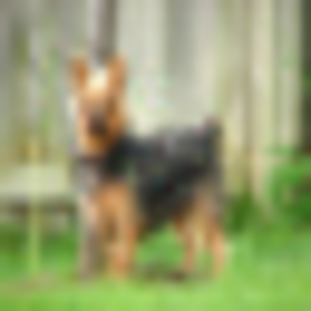}
         \end{subfigure}
         }
\caption{Demonstration using MNIST~\citep{lecun1998mnist} and CIFAR10~\citep{krizhevsky2009learning} imagery. (a,g) Natural images (labelled digit ``4'' and ``dog'' respectively); (b,h) Adv perturbations generated by PGD; (c,i) Adv images by applying the perturbations (the predictions of the classifier are changed to digit ``9'' and ``deer'' respectively);
(d, j) the vulnerability heatmaps learned by our method (``hotter'' means more vulnerable); (e,k) Selected perturbations of PGD according to the vulnerability heatmaps; (f,l) Refined PGD adv images (the predictions of the classifier are the same as the original PGD). Images have been resized to higher resolutions for better visualisation.}
\label{fig-demo}
\vspace{-0.5cm}
\end{figure}

In this paper, we would like to tackle the task of generating adv images with a fewer number of perturbations from a different perspective than existing sparse attacks. 
Our general idea is as follows: \textit{noting that dense attacks generate dense yet small perturbations, given an adv image with the perturbations generated by a dense attack, can we significantly reduce the number of perturbations by selecting the most effective ones and keep the attack performance as per the original adv image?}
As dense attacks are usually much faster than sparse ones, with an efficient selection process, we can devise a new framework that attacks with fewer perturbations more efficiently. 
As distinct from sparse attacks, which use independent methods for integrating the search for perturbation locations and magnitudes, ours is a post-hoc refinement framework for dense attacks, which decomposes the search for magnitude and location to the input dense attacks and the refinement process, respectively.

To further motivate our idea, we demonstrate that a large proportion of the generated perturbations of many dense attacks contribute little to the attack against the classifier.
For example, in Figure~\ref{fig-demo} (b,h), we can see that the perturbations generated by Projected Gradient Descent (PGD)~\citep{madry2017towards}) are distributed across all the pixels of the natural images. However, shown in Figure~\ref{fig-demo} (e, k, f, l), we can fool the classifier's predictions similarly to PGD, with only 10\% of its perturbations.
The key research question is how to identify the vulnerable pixel locations of an adv image generated by a dense attack. Here we propose to identify an image's \textit{pixel vulnerability} of the target classifier, where perturbations imposed on the more vulnerable pixels are more likely to change the prediction of the classifier. Figure~\ref{fig-demo}(d,j) show the pixel vulnerability heatmaps (we use the words ``heatmap'' and ``map'' interchangeably) discovered by our approach from PGD, with which we can choose the perturbations accordingly.
This brings two important benefits: \textbf{1)} With the reduced number of perturbed pixels, we can not only improve the imperceptibility of dense attacks but also significantly lower the accuracy of adv detection methods~\citep{feinman2017detecting,xu2017feature,ma2018characterizing,pang2018towards,ma2020understanding} on them.
\textbf{2)} The vulnerability heatmap can greatly help us to get a more intuitive and deeper understanding of the semantic structures of dense attacks.
For example, Figure~\ref{fig-demo}(d) highlights how the digit ``4'' is changed to digit ``9'', which concurs with our visual perception, and in Figure~\ref{fig-demo}(j) where PGD tries to add perturbations to generate a pair of ``fake'' antlers, which fools the classifier to think it as an image of a ``deer''.

The contributions of our paper can be summarised as follows:
\textbf{1)} We tackle the task of attacking with fewer perturbations with a novel approach, that refines given dense attacks in a post-hoc manner and is different from other sparse attacks. To the best of our knowledge, our methodology is original.
\textbf{2)} Given an adv image of a dense attack, we propose to learn its vulnerability map with a DNN, which is used to select the most vulnerable pixels to attack so as to preserve the dense attack's performance but with significantly fewer perturbations. As the proposed framework is trained with adv images generated by dense attacks as input, it can be used to refine an arbitrary dense attack.
\textbf{3)} In contrast to input dense attacks, our framework can generally reduce to 70\% of the number of perturbations whilst keeping their attack performance as well as significantly reducing their detectability by both human and adversarial detectors. \textbf{4)} Compared with sparse attacks, our method can significantly improve the attack speed with better performance. 

\begin{figure*}
    \centering
    \resizebox{0.86\textwidth}{!}{
    \includegraphics{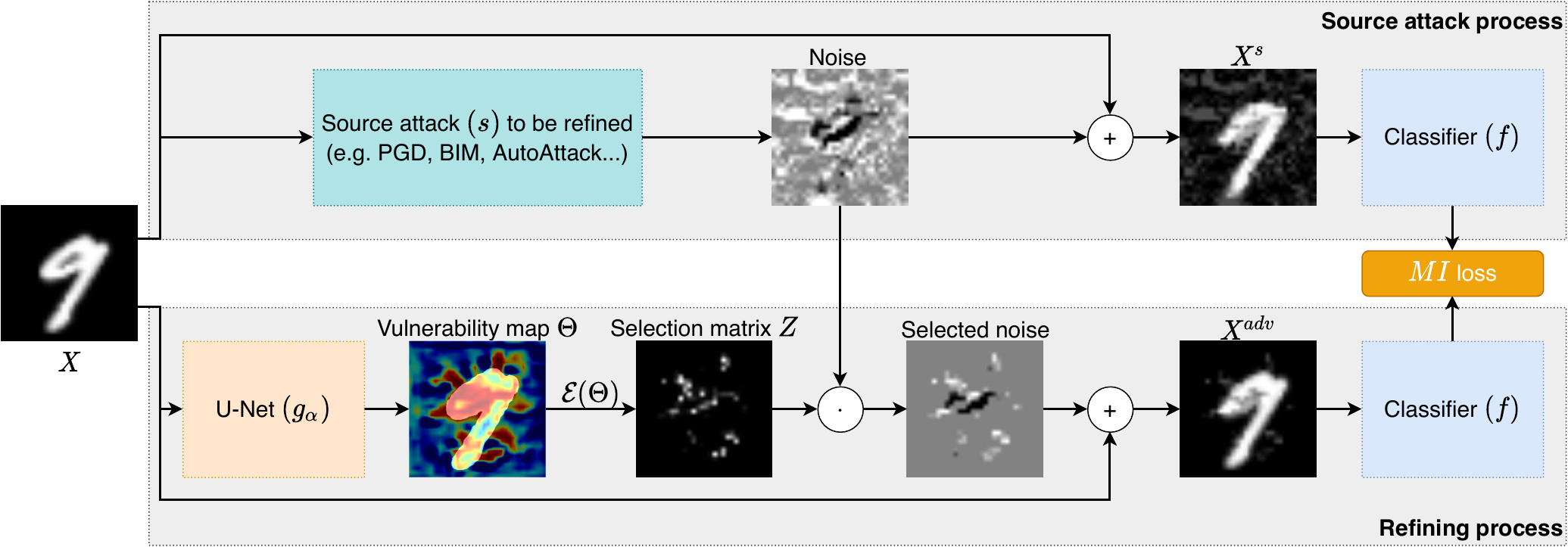}
    }
    \caption{Overview of our framework. Our method is designed to refine a given source (dense) attack by: 1) invoking the source attack to generate adversarial perturbations/noise, i.e., $\matr{X^{s}} - \matr{X}$; 2) taking the natural image $\matr{X}$ as input to generate the vulnerability map $\matr{\Theta}$; 3) generating the selection matrix $\matr{Z}$ according to $\matr{\Theta}$; 4) selecting a subset of the perturbations/noise generated by the source attack; 5) generating a new adversarial image $\matr{X^{adv}}$  with the selected perturbations.} 
    \label{fig-arch}
    \vspace{-0.5cm}
\end{figure*}

\section{PROPOSED APPROACH}
\subsection{Background and Problem Definition}
\label{sec-background}
Suppose that a normal image and its ground-truth label are denoted by $\matr{X}$ and $\hat{y} \in \{1,\cdots,K\}$, respectively, where $K$ is the number of unique image labels. We consider a pretrained neural network classifier $f$, which takes $\matr{X}$ as input and outputs a probability vector over $K$ labels, i.e., $\vec{y}=f(\matr{X})$. Usually for a well-trained classifier, we have $\hat{y}=\argmax \vec{y}$. 
An adversarial attack is to find perturbations $\matr{\eta}$ (with the same dimensions as $\matr{X}$) and add them to $\matr{X}$ to give the adv example, $\matr{X}^{adv}$, which is expected to mislead the classifier to fail to identify the true label.
Importantly, the adv image shall be inside an $\epsilon$-ball around $\matr{X}$, i.e., $\Vert \eta \Vert < \epsilon$ where $\Vert \bcdot \Vert$ can be the $L_{\infty}$ norm in accordance with~\cite{madry2017towards,athalye2018obfuscated} or other norms. Finding the value of $\eta$ can be formulated as the following optimisation problem:
$\max_{\Vert \eta \Vert < \epsilon} \ell(f(\matr{X} + \eta), \hat{y}),$
where $\ell(f(\matr{X}), \hat{y})$ denotes the loss function (e.g., cross-entropy loss) of the classifier $f$ given input $\matr{X}$ and its true label $\hat{y}$.

Recall that our goal is to refine the adv images of a given dense attack, which we call the \textit{source attack}.
We assume an image $\matr{X}$ consists of $H \times W$ pixels, where each pixel $x_{ij}$ can have either one channel (greyscale images) or multiple channels (colour images)\footnote{We view a multi-channel image as a matrix as we are only interested in the pixel coordinates in lieu of  the values of the pixels.}.
Given $\matr{X}$ and a target classifier $f$, we can apply the source attack $s$
to generate the source adversarial image, $\matr{X}^{s}$: $\matr{X}^{s} = s(\matr{X})$, whose predicted label distribution is $\vec{y}^s = f(\matr{X}^s)$.
We also assume that the source attack can challenge the classifier well, i.e., $\argmax \vec{y^s} \neq \argmax \vec{y}$ and its perturbations fall into the $\epsilon$-ball.
Note that we only need the output of the source attack and need no knowledge of its implementation details.

Next, we introduce a vulnerability map $\matr{\Theta} \in \mathbb{R}^{H \times W}$, generated from a neural network $g$ parameterised by $\alpha$ taking $\matr{X}$ as input: $\matr{\Theta} = g_\alpha(\matr{X})$.
Specifically, $\theta_{ij}$ indicates the vulnerability of pixel $x_{ij}$.
Based on $\matr{\Theta}$, we would like to develop a procedure $\mathcal{E}$ (detailed in Section~\ref{sec-sel}) that discretely selects a subset of the most vulnerable pixels to perturb. 
Specifically, a draw from $\mathcal{E}$ is a binary selection matrix: $\matr{Z} \sim \mathcal{E}(\matr{\Theta}) \in \{0,1\}^{H \times W}$, where $z_{ij}=1$ indicates pixel $x_{ij}$ is selected and vice versa. Given $\matr{X}^{s}$ and $\matr{Z}$, we construct a new adv example by imposing the source attack's perturbations only on the selected pixels:
\begin{equation}
\label{eq-l2a-adv}
\matr{X}^{adv} = \matr{X} + \matr{Z} \odot (\matr{X}^{s} - \matr{X}),
\vspace{-3mm}
\end{equation}
where $\odot$ indicates the element-wise product and $\eta = \matr{X}^{s} - \matr{X}$ consists of the perturbations from the source attack.
In this way, $\matr{X}^{adv}$ is constructed by imposing a subset of perturbations from the source adv image $\matr{X}^s$ on the natural image $\matr{X}$.
Finally, we can describe our research problem as: \textit{Given $\matr{X}$ and $\matr{X}^{s}$, finding the optimal $\matr{\Theta}$ (i.e., $\matr{Z}$) to generate a new adversarial image $\matr{X}^{adv}$ in the $\epsilon$-ball, which can attack as well as the source adversarial image $\matr{X}^s$}.
We illustrate the research question and our proposed method in Figure~\ref{fig-arch}. More details of our proposed framework will be elaborated on in the following sections.

\subsection{Training Objective}
\label{sec-learning}

Given the problem definition, we introduce our learning algorithm derived from an information-theoretic perspective.
Recall that our objective is to let $\matr{X}^{adv}$ challenge the classifier $f$ similarly to $\matr{X}^s$.
From the probabilistic perspective, if we view $\matr{X}^{adv}$ as a random variable parametrised by $\matr{\Theta}$, then the objective can be achieved by learning $\matr{\Theta}$ to push $\matr{X}^{adv}$ close to $\vec{y}^s$ (the  predicted label distribution of the source attack).
The mutual information (MI) is a widely-used measure of dependence between two random variables, which captures how much knowledge of one random vector reduces the uncertainty about the other~\citep{cover1999elements}.
Therefore, we can formulate the above problem as the following maximisation of the mutual information:
\begin{equation}
\label{eq-l2a-obj-mi}
\mathcal{L} = \max_{\matr{Z}\sim \mathcal{E}(\matr{\Theta})} \mathbb{I}\left(\vec{y}^s, \matr{X}^{adv}\right).
\vspace{-3mm}
\end{equation}
As $\matr{X}^{s}$ comes from the source attack, it already satisfies the constraint of the $\epsilon$-ball. Therefore, there is no need to consider this constraint when learning $\matr{Z}$.
We show that Eq.~(\ref{eq-l2a-obj-mi}) can be written as (the details are in the supplementary material):
\begin{align}
\label{eq-mi-derive}
\exptt{\matr{X}}{\exptt{\matr{X}^{s}, \matr{Z}}{\exptt{\matr{X}^{adv}}{\sum_{k=1}^{K} p(y_k|\matr{X}^s) \log p(y_k|\matr{X}^{adv})}}}.
\vspace{-4mm}
\end{align}
Specifically, we have: 1) $\matr{X}^{s}|\matr{X} \eqdef s(\matr{X})$, which is the process of generating adversarial images from the source attack; 2) $\matr{Z} | \matr{X} \eqdef \matr{Z} \sim \mathcal{E}(\matr{\Theta}), \matr{\Theta} = g_\alpha(\matr{X})$, which is the proposed process of generating the vulnerability map and selection matrix; 3) $\matr{X}^{adv}|(\matr{X}^{s}, \matr{Z}) \eqdef \matr{X} + \matr{Z} \odot (\matr{X}^{s} - \matr{X})$, which is the process of imposing the selected perturbations on the natural image;
4) $p(y_k|\matr{X}^s) \propto f(\matr{X}^s)_{k}$, and $p(y_k|\matr{X}^{adv}) \propto f(\matr{X}^{adv})_{k}$, which correspond to the prediction processes of the source and our new adv images, respectively.
To summarise, the learning process of our refinement framework can be described as: for one input image $\matr{X}$, we first apply the source attack to get $\matr{X}^s$, we then sample the selection matrix $\matr{Z}$; then we accordingly generate $\matr{X}^{adv}$; $\matr{X}^{s}$ and $\matr{X}^{adv}$ are fed into the classifier to get their predicted label distributions; finally, we minimise the cross-entropy between the two label distributions.

\subsection{Construction of Selection Process}
\label{sec-sel}
Now we introduce the details of the process $\mathcal{E}$ that generates $\matr{Z}$. Inspired by~\cite{chen2018learning}, among all the $HW$ pixels of an image, we maximally select $M = \lceil \beta H W \rceil$ pixels to impose on the perturbations from the source attack where $\beta \in (0,1)$ is the proportion of the maximally selected pixels. Next, we introduce a matrix version of the categorical distribution parameterised by $\matr{\Theta}$, a draw of which picks one pixel among the $HW$ pixels: $\matr{U} \sim \text{Categorical}(\matr{\Theta})$.
Here $\matr{U} \in \{0,1\}^{H \times W}$ is a one-hot binary matrix, where the entry selected by the categorical distribution is turned on and all the others are zeros\footnote{An ordinary categorical distribution's parameter is a probability vector and it outputs a one-hot vector. A more precise notation should be $\matr{U} \sim \text{reshape\_to\_matrix}(\text{Categorical}(\text{softmax}(\text{reshape\_to\_vector}(\matr{\Theta}))$. We abbreviate it to assist the readability.}.

To maximally select $M$ pixels, we draw $\matr{U}$ $M$ times and choose the pixels that are selected at least once:
\begin{align}
\label{eq-r}
\matr{U}^{<m>} &\sim \text{Categorical}(\matr{\Theta})\text{~~~for $m=1$ to $M$},\nonumber\\
\matr{Z} &= \max_{m} \matr{U}^{<m>},
\vspace{-3mm}
\end{align}
where $\matr{U}^{<m>}$ denotes the output matrix of the $m^{\text{th}}$ draw and $\max_m \matr{U}^{<m>}$ denotes the element-wise maximisation, i.e., $z_{ij} = \max_m u^{<m>}_{ij}$.

Recall that we aim to train the neural network parameter $\alpha$ to optimise Eq.~(\ref{eq-l2a-obj-mi}). To use backpropagation through the categorical random variables, we utilise the Concrete distribution~\citep{jang2017categorical,maddison2016concrete} to approximate the categorical distribution: $\matr{\tilde{U}} \sim \text{Concrete}(\log \matr{\Theta})$, where $\matr{\tilde{U}} \in \mathbb{R}_+^{H \times W}$ is the continuous relaxation of the one-hot matrix  $\matr{U}$ in Eq.~(\ref{eq-r}). Specifically, we have:
\begin{equation}
\tilde{u}_{ij} = \frac{\exp\left(\log (\theta_{ij} + \gamma_{ij})/\tau\right)}{\sum_{i'j'}^{H\times W} \exp\left(\log (\theta_{i'j'} + \gamma_{i'j'})/\tau\right)},
\end{equation}
where $\gamma_{ij}$ is from a $\text{Gumbel}(0,1)$ distribution: $\gamma_{ij} = -\log(-\log \nu_{ij}), \nu_{ij} \sim \text{Uniform}(0,1)$, and $\tau$ is the ``temperature''. 
With the Concrete distribution, we have:
\begin{align}
\label{eq-r-prime}
\matr{\tilde{U}}^{<m>} &\sim \text{Concrete}(\log \matr{\Theta}) \text{~~~for $m=1$ to $M$},\\
\label{eq-z}
\tilde{\matr{Z}} &= \max_m \matr{\tilde{U}}^{<m>},
\end{align}
where $\matr{\tilde{Z}}$ is an approximation to $\matr{Z}$ for backpropogation in the learning phase.
\begin{algorithm}
\SetKwInOut{Input}{input}\SetKwInOut{Output}{output}
\Input{Classifier $f$, To-be-refined source attack $s$, Input image collection $\mathcal{X}$, \\ Max proportion of vulnerable pixels $\beta$}
\Output{Parameter of the neural network $\alpha$}
\While{Not converged}
{
	Sample a batch of input images $\mathcal{X}_B$\;
	\ForAll{$\matr{X} \in \mathcal{X}_B$}
	{
		Generate source adversarial image $\matr{X}^s=s(\matr{X})$\;
		\For{$m=1$ to $M$}
			{
			Draw $\matr{\tilde{U}}^{<m>}$ by Eq.~(\ref{eq-r-prime})\;
			}
		Calculate $\matr{\tilde{Z}}$ by Eq.~(\ref{eq-z})\;
		Generate final adversarial image $\matr{X}^{adv} = \matr{X} + \matr{\tilde{Z}} \odot (\matr{X}^{s} - \matr{X})$\;
		Calculate learning loss $\mathcal{L}$ by Eq.~(\ref{eq-l2a-obj-mi})\;
		Compute gradients of $\mathcal{L}$ in terms of $\alpha$\;
	}
	Average noisy gradients of batch samples\;
	Update $\alpha$ by stochastic gradient steps\;
}
\caption{Learning algorithm}
\label{alg}
\end{algorithm}

\subsection{Implementation and Learning Algorithm}
Here we give a further introduction to the implementation of our framework. Specifically, the only learnable component is the neural network $g_\alpha(\cdot)$, which takes $\matr{X}$ as input and outputs the vulnerability map $\matr{\Theta}$. In general, there is no limitation of implementation of $g_\alpha(\cdot)$. As $\matr{\Theta}$ has the same dimensions with $\matr{X}$, inspired by the rich works in image segmentation and saliency detection, we empirically find that U-Net~\citep{ronneberger2015u} yields good performance.
Finally, we elaborate on the training process in Algorithm~\ref{alg}. After the neural network $g$ is trained, given an input image $\matr{X}$, we first get $\matr{\Theta}$, then select from the perturbations of the source attack, and finally use Eq.~(\ref{eq-l2a-adv}) to generate the refined adv image $\matr{X}^{adv}$.

\section{RELATED WORK}
\textbf{Dense Attacks.~}
Dense attacks are the most popular category of adv attacks, where all the pixels of a natural image are perturbed, such as in FGSM~\citep{goodfellow2014explaining}), PGD~\citep{madry2017towards}), BIM~\citep{kurakin2016adversarial}, DeepFool~\citep{moosavi2016deepfool}, CW~\citep{carlini2017towards}, AutoAttack~\citep{croce2020reliable}.
Due to the space limit, a comprehensive review of dense attacks is beyond the focus of this paper. Theoretically, any dense attacks can serve as our source attacks and be refined by our proposed approach, as we only take the generated adversarial images as input for training and attacking.

\textbf{Sparse Attacks.}
Existing sparse attacks can be categorised into black-box attacks (such as~\cite{su2019one,narodytska2016simple,schott2018towards,croce2019sparse}) and white-box attacks, where the latter is more relevant to our approach.
As an example of white-box sparse attacks, \cite{carlini2017towards,zhao2018admm,modas2019sparsefool} propose variants of dense attacks that integrate the $l_0$ constraint to encourage sparsity.
Moreover, the Jacobian-based Saliency Map Attack (JSMA)~\citep{papernot2016limitations} uses the saliency map to determine the pixel positions to perturb, where perturbations are imposed on the pixels with high saliency. More recently, \cite{sapfECCV2020} proposes to factorise adversarial perturbations into two factors, which capture the magnitudes and positions, respectively.

\textbf{Fundamental Differences between Our Attacks and Sparse Attacks.}
At the concept level, though both our method and existing sparse attacks  
fall into the general context of ``reducing the number of perturbed pixels for adversarial attacks'', \textit{the thinking and methodology of our work are completely different}. 
Firstly, a sparse attack is an independent attack and many existing sparse attacks are achieved by modifying the inner steps of a dense attack (e.g., adding $l_0$ regularisation to the CW~\citep{carlini2017towards}), while our method is a post-hoc framework that refines dense attacks. As ours does not need to modify the inner steps of dense attacks but uses their output, it can be used to refine an arbitrary dense attack.
Secondly, the fundamental difference in the design and thinking results in several unique properties and advantages of our method: \textbf{1)} Sparse attacks focus more on perturbation positions than magnitudes, thus, resulting in fewer but larger perturbations, while ours selects more but much smaller perturbations,
which leads to more natural-looking adversarial images (see Figure~\ref{fig-vis}(a,b)).
\textbf{2)} Most sparse attacks do complex optimisation for searching the perturbations locations, whiles ours is based on neural networks and has much better efficiency (see Table~\ref{tb-sparse}). 
\textbf{3)}  With vulnerability maps, our method helps to gain a better and more intuitive understanding of the semantic structures of the source attacks and of classifier vulnerability that further assists the development of robust models (see Fig.~\ref{fig-vis}(c)).

\section{EXPERIMENTS}\label{sec-experiments}
In this section, we conduct comprehensive experiments to evaluate the performance of our framework, named \textbf{PVAR} (\textbf{P}ixel \textbf{V}ulnerability \textbf{Ad}versary \textbf{R}efinement), on attacking image classifiers.
Here we focus on untargeted white-box attacks, though our approach has the potential to be adapted to targeted or black-box settings.

\begin{figure*}[t]
        \centering
         \begin{subfigure}[b]{0.24\linewidth}
                 \centering
                 \caption{BIM}
                 \includegraphics[width=0.99\textwidth]{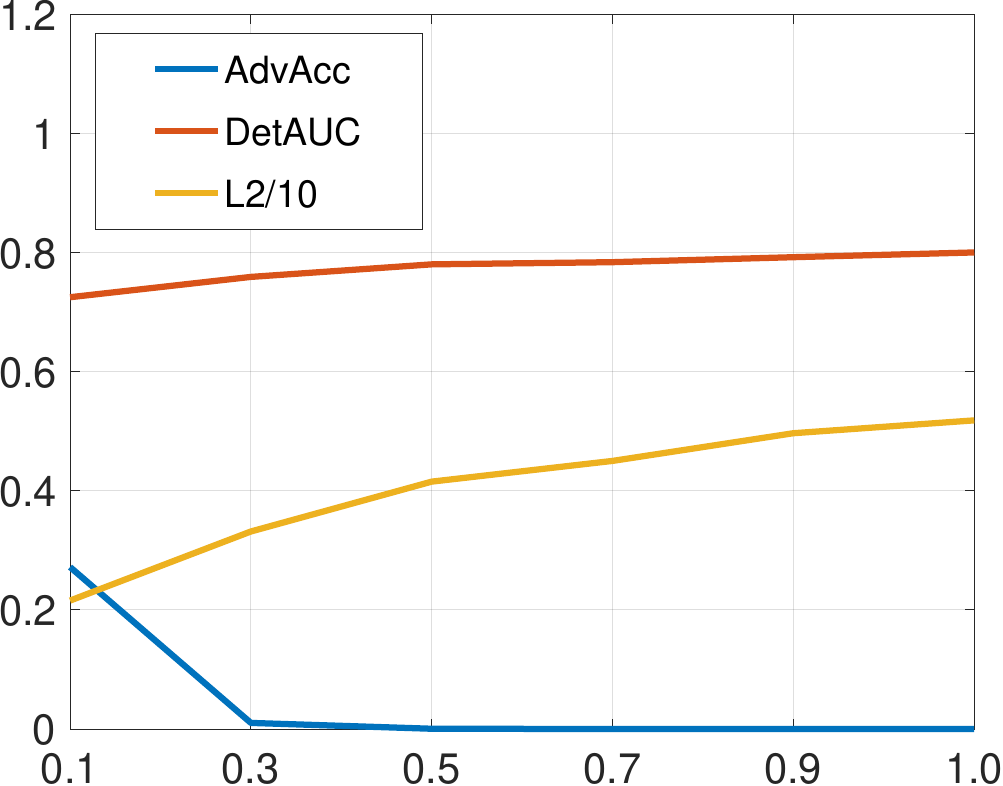}
         \end{subfigure}
         \begin{subfigure}[b]{0.24\linewidth}
                 \centering
                 \caption{PGD}
                 \includegraphics[width=0.99\textwidth]{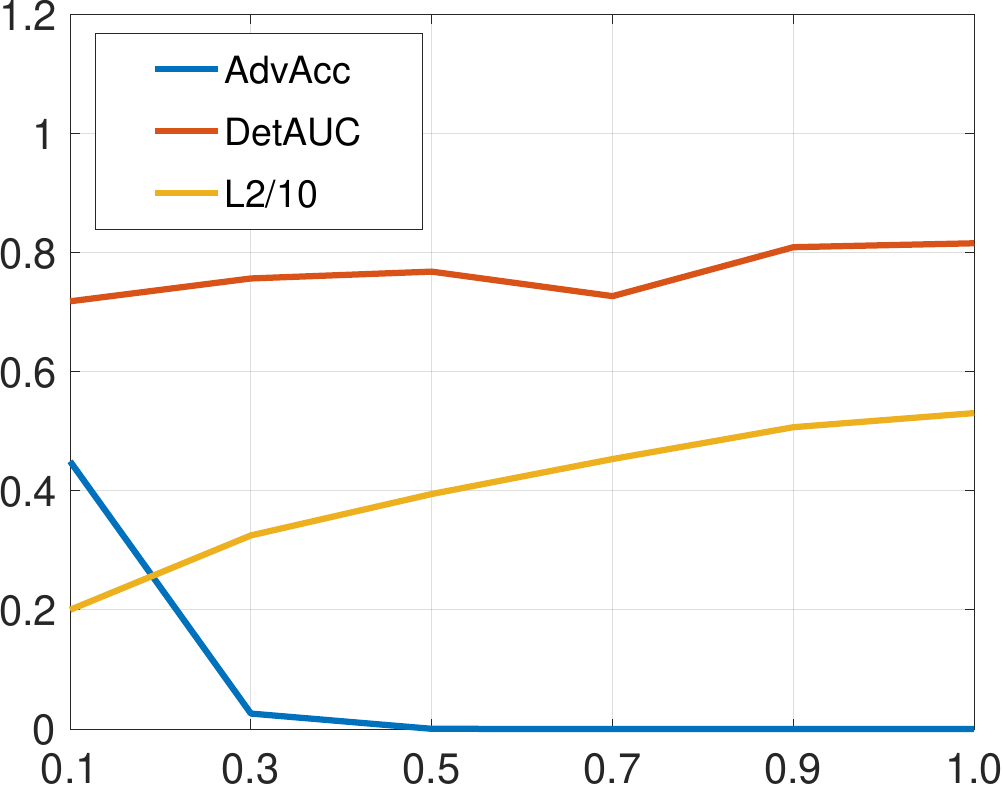}
         \end{subfigure} 
         \begin{subfigure}[b]{0.24\linewidth}
                 \centering
                 \caption{AutoAttack}
                 \includegraphics[width=0.99\textwidth]{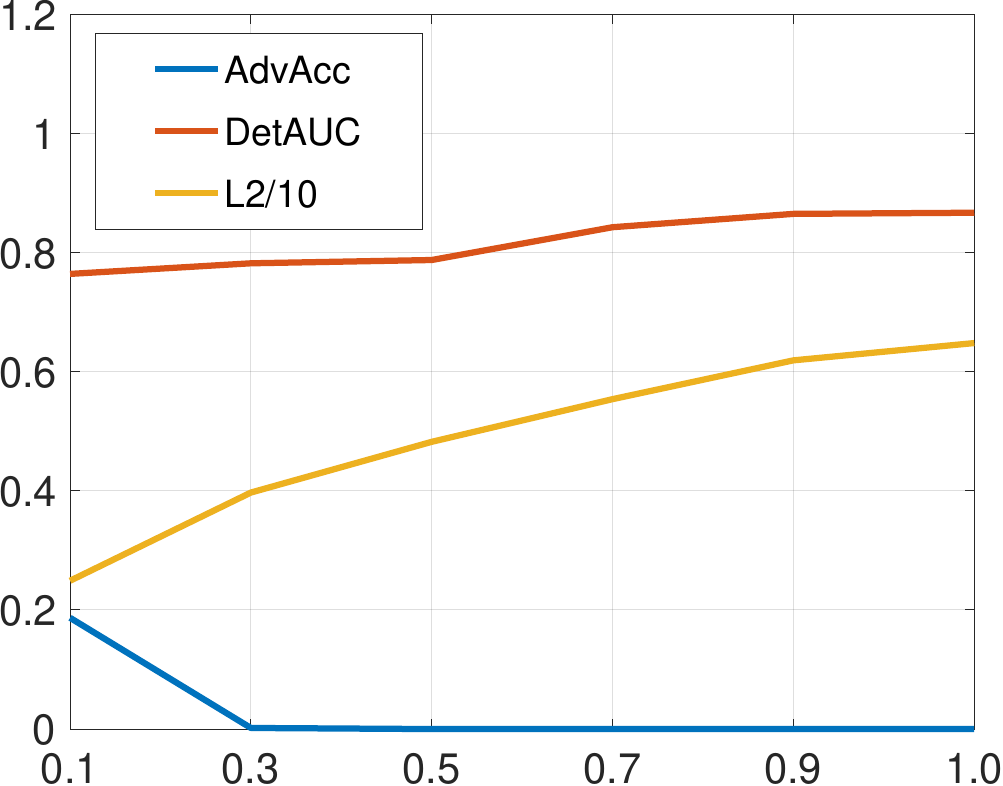}
         \end{subfigure}%
         \vspace{0.3cm}
         \caption{Refinement of various source attacks by PVAR on MNIST with LeNet as the classifier. The horizontal axis indicates the value of $\beta$. When $\beta=1.0$, it shows the performance of the source attack without any refinement, i.e., all the perturbations are applied. 
         ``L2/10'' means that we divide $l_2$ by 10 to show it in the similar range with DetAUC and AdvAcc.}
         \label{fig-mnist}
\end{figure*}

\begin{figure*}[t]
\begin{minipage}{0.7\textwidth}
        \centering
         \begin{subfigure}[b]{0.32\linewidth}
                 \centering
                 \caption{BIM}
                 \includegraphics[width=0.99\textwidth]{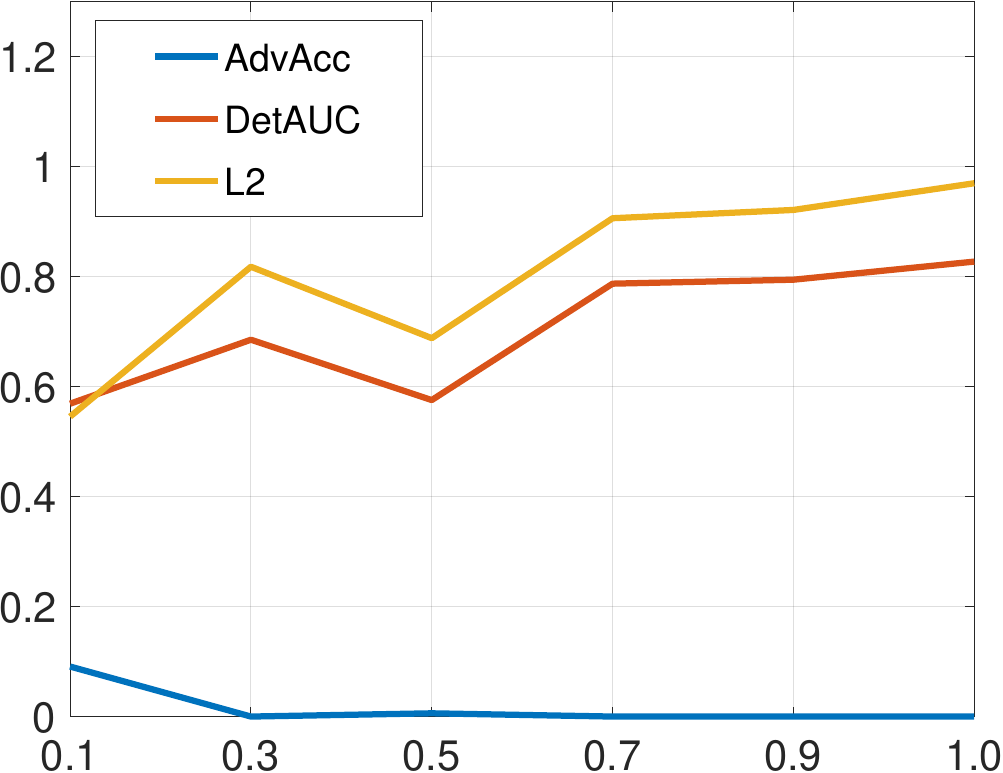}
         \end{subfigure}
         \begin{subfigure}[b]{0.32\linewidth}
                 \centering
                 \caption{PGD}
                 \includegraphics[width=0.99\textwidth]{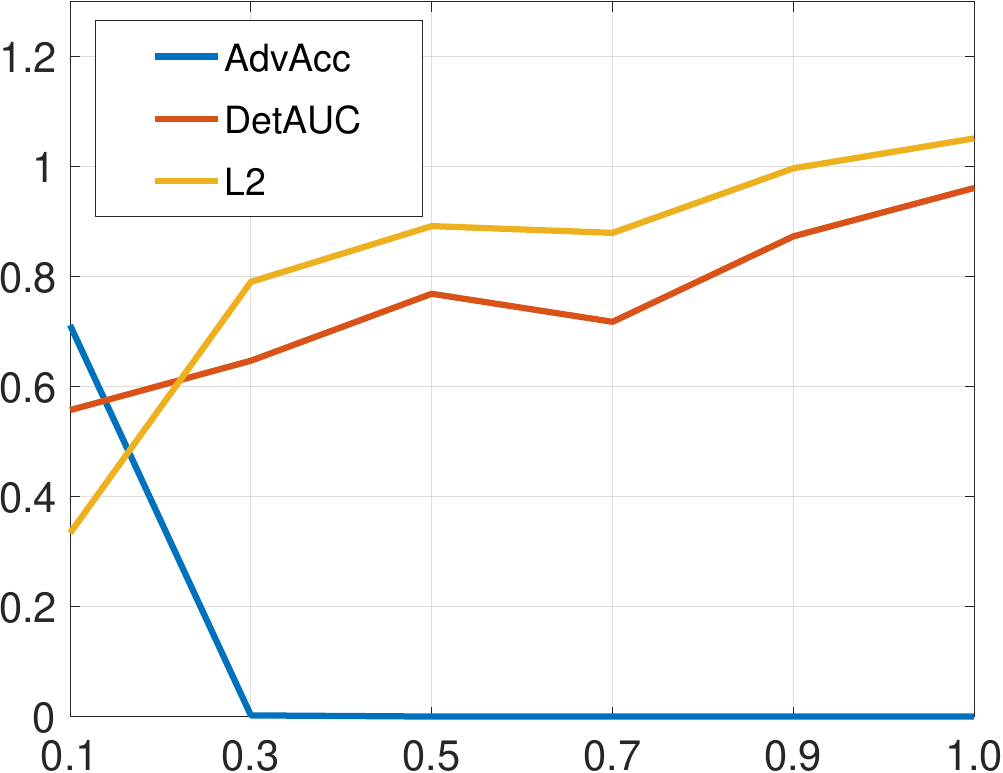}
         \end{subfigure} 
         \begin{subfigure}[b]{0.32\linewidth}
                 \centering
                 \caption{AutoAttack}
                 \includegraphics[width=0.99\textwidth]{figs/resnet32_pgd-crop.pdf}
         \end{subfigure}%
            \\
               \centering
         \begin{subfigure}[b]{0.32\linewidth}
                 \centering
                  \caption{BIM}
                 \includegraphics[width=0.99\textwidth]{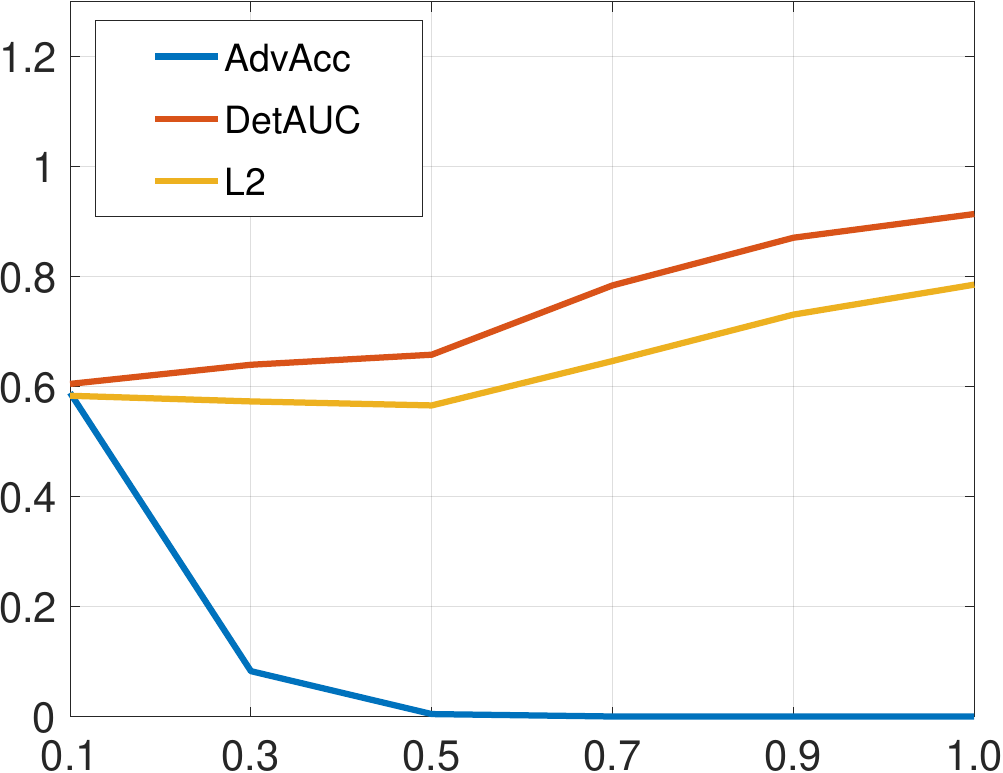}
         \end{subfigure}
         \begin{subfigure}[b]{0.32\linewidth}
                 \centering
                 \caption{PGD}
                 \includegraphics[width=0.99\textwidth]{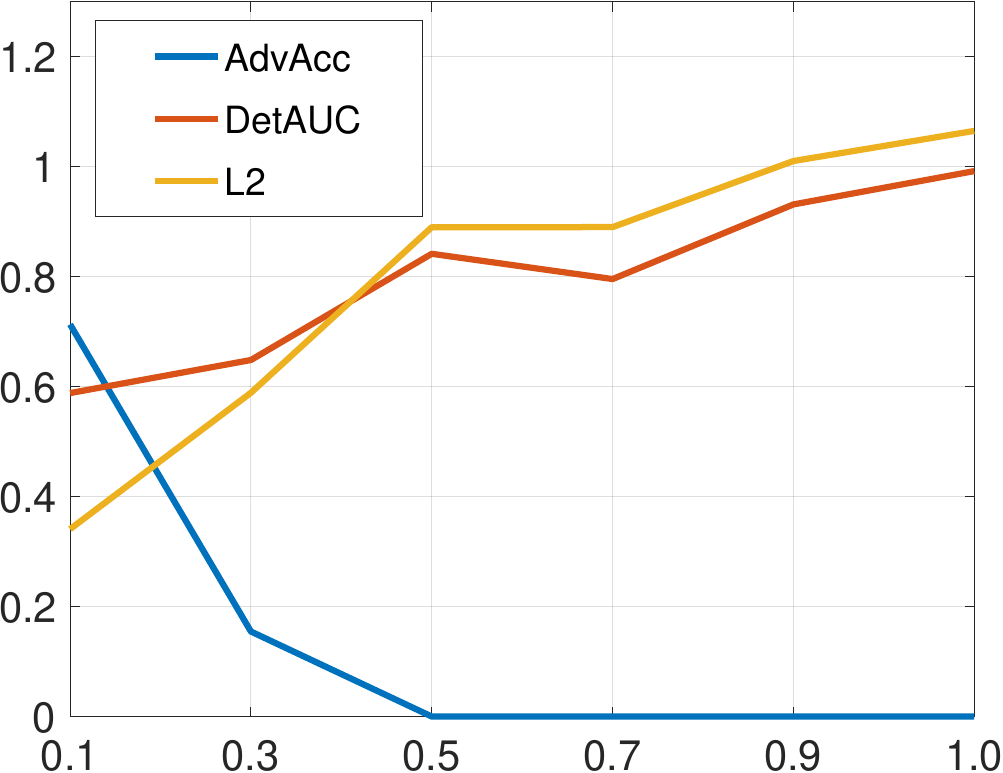}
         \end{subfigure} 
         \begin{subfigure}[b]{0.32\linewidth}
                 \centering
                 \caption{AutoAttack}
                 \includegraphics[width=0.99\textwidth]{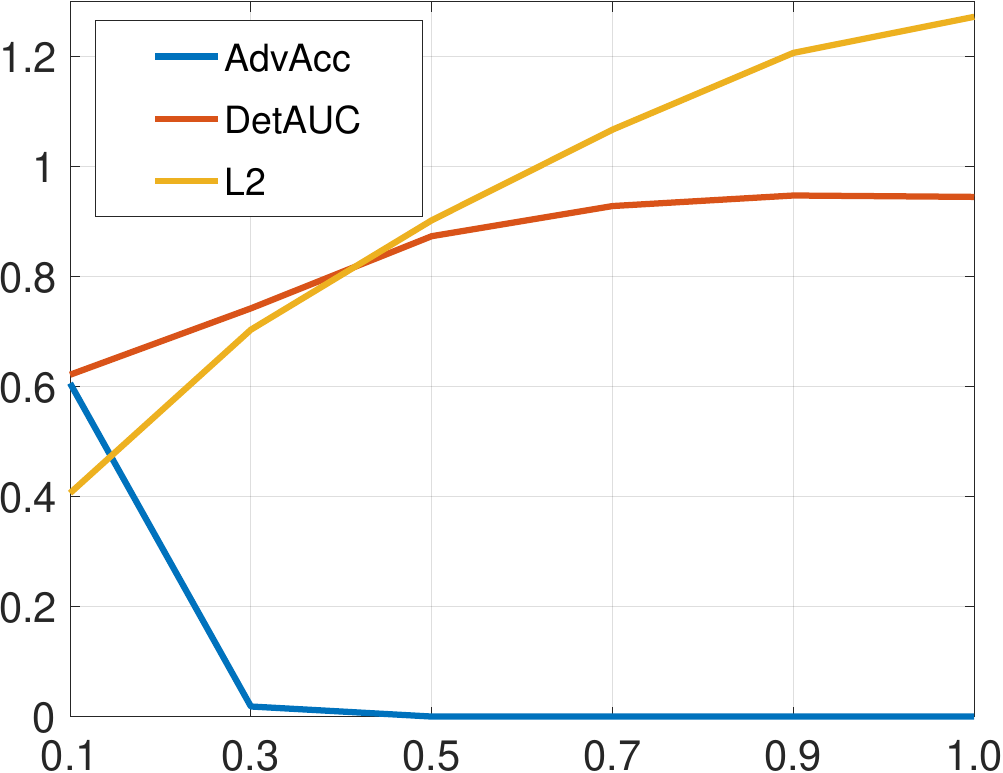}
         \end{subfigure}%
                  \caption{Refinement of various source attacks by PVAR on CIFAR10. First row: ResNet32 as the classifier. Second row: ResNet56 as the classifier. The meaning of $\beta$ is the same as in Figure~\ref{fig-mnist}.}
        \label{fig-cifar}
\end{minipage}
\hspace{0.005\textwidth}
\begin{minipage}{0.29\textwidth}
    \centering
    \includegraphics[width=1.0\linewidth]{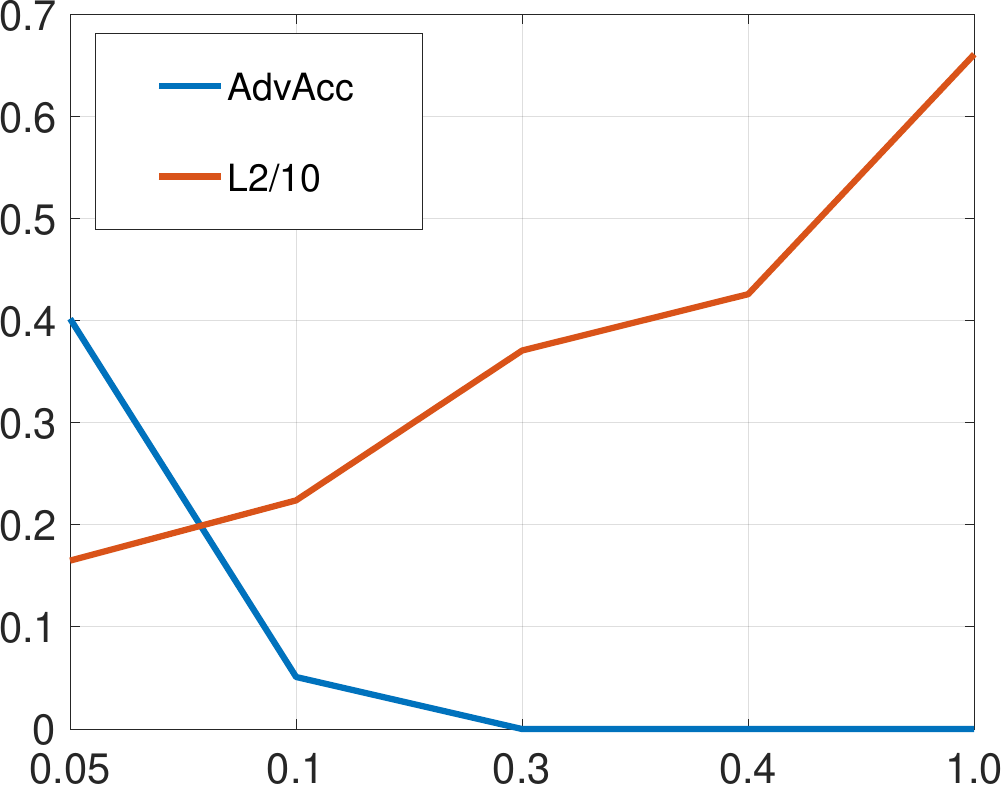}
    \caption{Refinement of BIM by PVAR on ImageNet with DenseNet169 as the classifier. 
    The meaning of $\beta$ is the same as in Figure~\ref{fig-mnist}.}
    \label{fig-in}
\end{minipage}
        \vspace{-0.6cm}
\end{figure*}

\textbf{Evaluation Metrics.~}
We use five sets of different metrics for comprehensive evaluations:
\textbf{1)} We use the classification accuracy on the adversarial images (\textbf{AdvAcc}) generated from the test images of a dataset, to test whether PVAR preserves the source attack's performance with fewer perturbations.
\textbf{2)} To verify the assumption that the refined attack is less easily detected than its source counterpart. We build an adversarial detector following~\cite{feinman2017detecting,pang2018towards} to predict whether an image is natural or adversarial. 
For an attack method, we split the test set of a dataset in half, then use the images of the first half and their adversarial counterparts to train the detector, and report the Area Under Curve (\textbf{DetAUC}) for the second half.
Note that a detector is trained individually for a specific attack (e.g., PGD, PVAR for PGD with $\beta=0.1$, and with $\beta=0.3$ are three different attacks and their detectors are trained separately).
The detector consists of a binary classifier with three advanced features carefully designed for adversarial detection: classifier confidence, kernel density (K-density), and the entropy of normalized non-maximal elements (non-ME). The former two features are introduced in~\cite{feinman2017detecting} and the latter is from~\cite{pang2018towards}.
As shown in Table 2 of~\cite{pang2018towards}, using non-ME only already achieves a very high DetAUC. Our detector is able to obtain stronger performance by combining multiple features.
\textbf{3)} To compare attack efficiency for each method, we also report the \textbf{averaged running time} of generating an adversarial image.
\textbf{4)} To measure \textit{how our refined attacks generate less perceivable adversarial images to human}, we report the averaged \textbf{$l_2$ distances} between them for each attack method that measure the closeness between natural images and their adversarial examples, i.e., a smaller the $l_2$ distance indicates that an adversarial example is closer to its natural counterpart, thus, it is less perceivable.
\textbf{5)} To qualitatively study ``how less perceivable'', we conduct a user study and report the \textbf{perceivableness} of attacks.
Finally, We note that from an attacker's perspective, \textit{the lower the five metrics are, the better.}

\textbf{Experimental Settings.~}
Here we consider the MNIST~\citep{lecun1998mnist}, 
CIFAR10~\citep{krizhevsky2009learning} and ImageNet~\citep{imagenet_cvpr09} datasets, where 
the pixel values of the images are normalised between 0 and 1.
\textbf{Settings of classifiers: } We use several widely-used pretrained covoluntional neural network models as the target classifiers:
LeNet~\citep{lecun1998gradient} for MNIST with 0.995 natural accuracy; ResNet32-v1~\citep{he2016deep} and ResNet56-v2~\citep{szegedy2016inception} for CIFAR10 with 0.932 and 0.921 accuracy,  respectively; DenseNet169~\citep{huang2017densely} for ImageNet reached 0.762 top-1 and	0.932 top-5 accuracy.
\textbf{Settings of source attacks: } We select three popular dense attacks as the example source attacks to be refined by our PVAR: PGD~\citep{madry2017towards}\footnote{\label{foolbox}Implemented in Foolbox~\citep{rauber2017foolbox}}, BIM~\citep{kurakin2016adversarial}\footref{foolbox} and AutoAttack~\citep{croce2020reliable}\footnote{Implemented in \url{https://github.com/fra31/auto-attack}}. 
We set the attack strength $\epsilon=0.3$ for MNIST, and $\epsilon=0.03$ for CIFAR10 and ImageNet. For other settings, we follow the standard ones.
Note that the source attack is not limited to the above three and can be an arbitrary dense attack.
\textbf{Settings of PVAR: } For the neural network $g$, we leverage a U-Net~\citep{ronneberger2015u} architecture, whose details are given in the supplementary material.
We vary the parameter that controls the proportion of the maximally selected pixels, $\beta$.
The temperature of the Concrete distribution is empirically set to 1.0.
We implement PVAR in Python with TensorFlow, trained by Rectified Adam~\citep{liu2019variance} with learning rate 0.001 and batch size 32. We train the model for 300 iterations maximally and terminate the training if the loss stops dropping for 15 continuous iterations. We use the datasets' standard training set to train PVAR.
\textbf{Settings of other baselines: } Besides the three source attacks that serve as natural baselines for the PVAR refined counterparts, we compare with several state-of-the-art sparse attacks including  CornerSearch~\citep{croce2019sparse} (a stronger baseline than CW-$l_0$~\citep{carlini2017towards})\footnote{\label{cs}Implemented in \url{https://github.com/fra31/sparse-imperceivable-attacks}}, JSMA~\citep{papernot2016limitations}\footnote{Implemented in \url{https://github.com/gongzhitaao/tensorflow-adversarial}}, and SparseFool~\citep{modas2019sparsefool} (a stonger baseline than the one-pixel attack~\citep{su2019one})\footref{foolbox}. We will specify the settings of the above sparse attacks in the next subsection.
\begin{table}[!ht]
\caption{Comparison with sparse attacks. We keep AdvAcc roughly the same for all the attacks and compare their DetAUC, $l_2$, and speed. Best results are in boldface.}
\label{tb-sparse}
\centering
\resizebox{0.9\linewidth}{!}{
\begin{tabular}{ccccc}
 \toprule
Attack                  & AdvAcc & DetAUC & $l_2$   & \begin{tabular}[c]{@{}c@{}}Attack Speed \\ (secs per image)\end{tabular} \\ \midrule \midrule
 & \multicolumn{4}{c}{MNIST (LeNet)} \\ \midrule
JSMA       & 0.007  & 0.88   & 7.21 &    0.03                                                                        \\ \midrule
SparseFool & 0.001  & 0.83   & 5.75 &  0.22                                                                        \\ \midrule
CornerSearch & 0.003  & 0.93   & \textbf{3.59} & 2.76 \\ \midrule 
\begin{tabular}[c]{@{}c@{}}Our refined \\ AutoAttack \end{tabular}& 0.001  & \textbf{0.78}   & 3.97 & \textbf{0.008}                                                                         \\ \midrule \midrule
 & \multicolumn{4}{c}{CIFAR10 (ResNet32)} \\\midrule
JSMA        & 0.002  & 0.95   & 1.59 &      0.18                                                                  \\ \midrule
SparseFool & 0.003  & 0.73   & 1.91 & 3.09                                                                         \\ \midrule
CornerSearch & 0.006  & 0.83   & 0.90 & 1.18 \\ \midrule
\begin{tabular}[c]{@{}c@{}}Our refined \\ BIM \end{tabular}           & 0.000  & \textbf{0.68}   & \textbf{0.81} & \textbf{0.02}                                                                      \\ \midrule \midrule
                        & \multicolumn{4}{c}{CIFAR10 (ResNet56)}                                                                     \\ \hline
JSMA       & 0.005  & 0.87   & 1.24 &        0.39                                                                  \\ \midrule
SparseFool & 0.003  & 0.76   & 5.53 &  10.41                                                                        \\ \midrule
CornerSearch & 0.002  & 0.79   & 2.01 & 1.92 \\ \midrule
\begin{tabular}[c]{@{}c@{}}Our refined \\ BIM  \end{tabular}          & 0.004  & \textbf{0.65}   & \textbf{0.57} & \textbf{0.03}                                                                         \\ \midrule \midrule
                        & \multicolumn{4}{c}{ImageNet (DenseNet169)} \\\hline  
JSMA       & 0.030  & -   & 8.30 &  365.86                                                                        \\ \midrule
SparseFool &  0.000 &  - & 9.15 & 32.92     \\ \midrule
\begin{tabular}[c]{@{}c@{}}Our refined \\ BIM \end{tabular}          & 0.000  & -   & \textbf{3.70} &  \textbf{0.06} \\\bottomrule
\end{tabular}
}
\vspace{-6mm}
\end{table}

\textbf{Results of Refinement of Dense Attacks.~}
We first show AdvAcc, DetAUC, and $l_2$ distance 
on MNIST, CIFAR10, and ImageNet in Figure~\ref{fig-mnist} and Figure~\ref{fig-cifar}, and Figure~\ref{fig-in}, respectively.
For MNIST and CIFAR10, we use all the test images, while for ImageNet, we randomly sample 1,000 images from the test set. As the detector~\citep{feinman2017detecting,pang2018towards} is originally proposed for MNIST and CIFAR10, and performs poorly on ImageNet, we omit its performance on ImageNet.
We divide $l_2$ on MNIST and ImageNet by 10 to show it in a similar range with DetAUC and AdvAcc.
We make the following remarks on the results:
\textbf{1)} It can be observed that all the dense attacks (i.e., $\beta=1.0$) achieve good attack performance, whose AdvAcc is zero.
However, they can be easily identified by the detector, shown in the high DetAUC scores. Especially for CIFAR10, the scores are generally greater than 0.95.
\textbf{2)} In most cases, our refinement framework can reduce at least 70\% (i.e., $\beta=0.3$) of the perturbations of the dense attacks but keep the AdvAcc nearly zero. This demonstrates that by wisely choosing the perturbations on the most vulnerable pixels, our framework can significantly reduce the number of perturbations without sacrificing the dense attacks' power. Moreover, PVAR shows its effectiveness in refining all the three attacks.
\textbf{3)} Reducing the number of perturbations brings us the obvious advantage of the refined attacks over the source ones in terms of $l_2$ and DetAUC, which correspond to adversarial attacks' detectability by human and detection algorithms, respectively.
For example, for AutoAttack on CIFAR10 with ResNet56, the refined AutoAttack with $\beta=0.3$ significantly reduces the original AutoAttack's $l_2$ from 1.27 to 0.70 and DetAUC from 0.94 to 0.74, while keeping the AdvAcc the same.

\begin{figure*}[!ht]
    \centering
         \begin{subfigure}[b]{0.34\linewidth}
                 \centering
                 \caption{}
                 \includegraphics[width=0.99\textwidth]{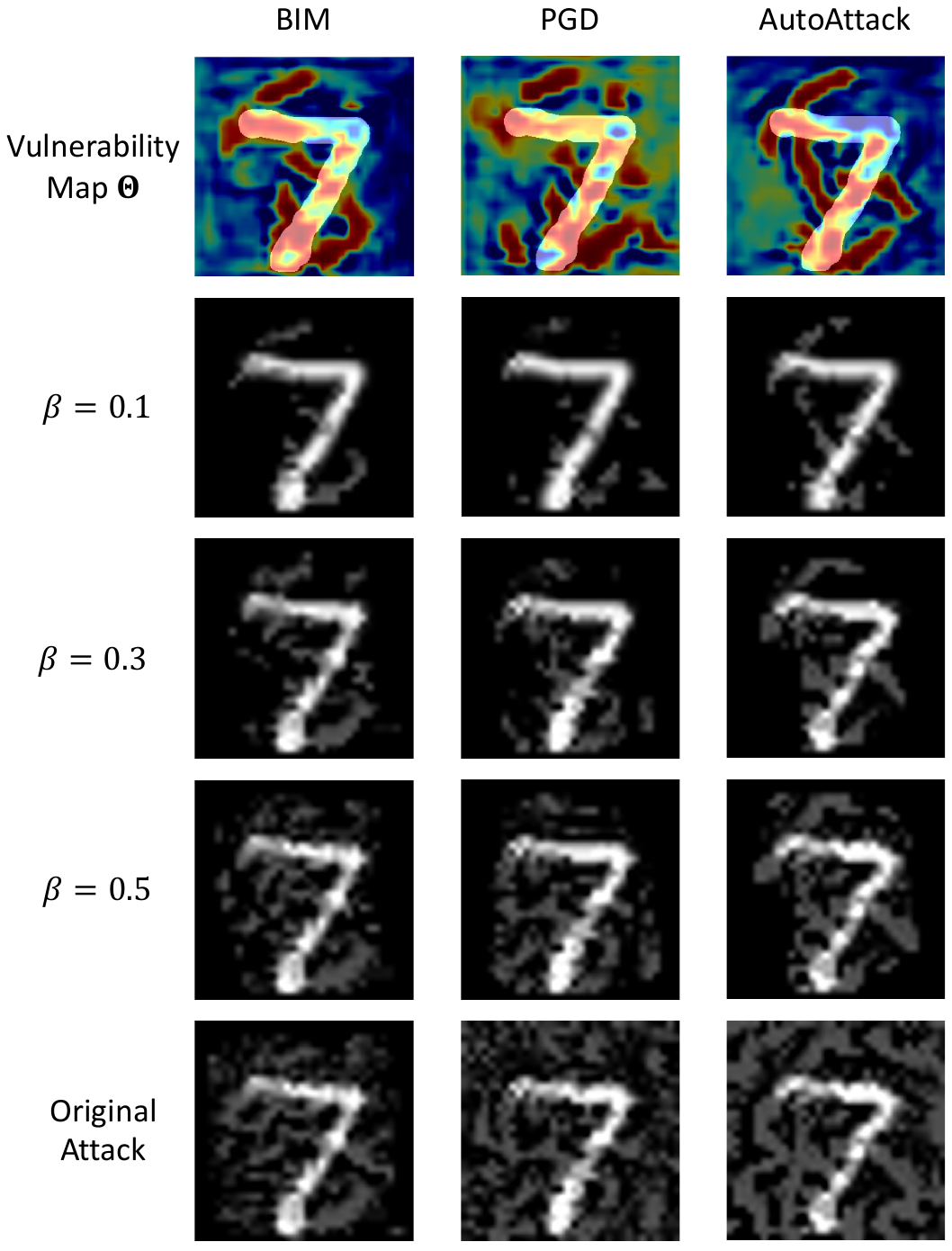}
         \end{subfigure}
         \begin{subfigure}[b]{0.62\linewidth}
                 \centering
                 \caption{}
                 \includegraphics[width=0.99\textwidth]{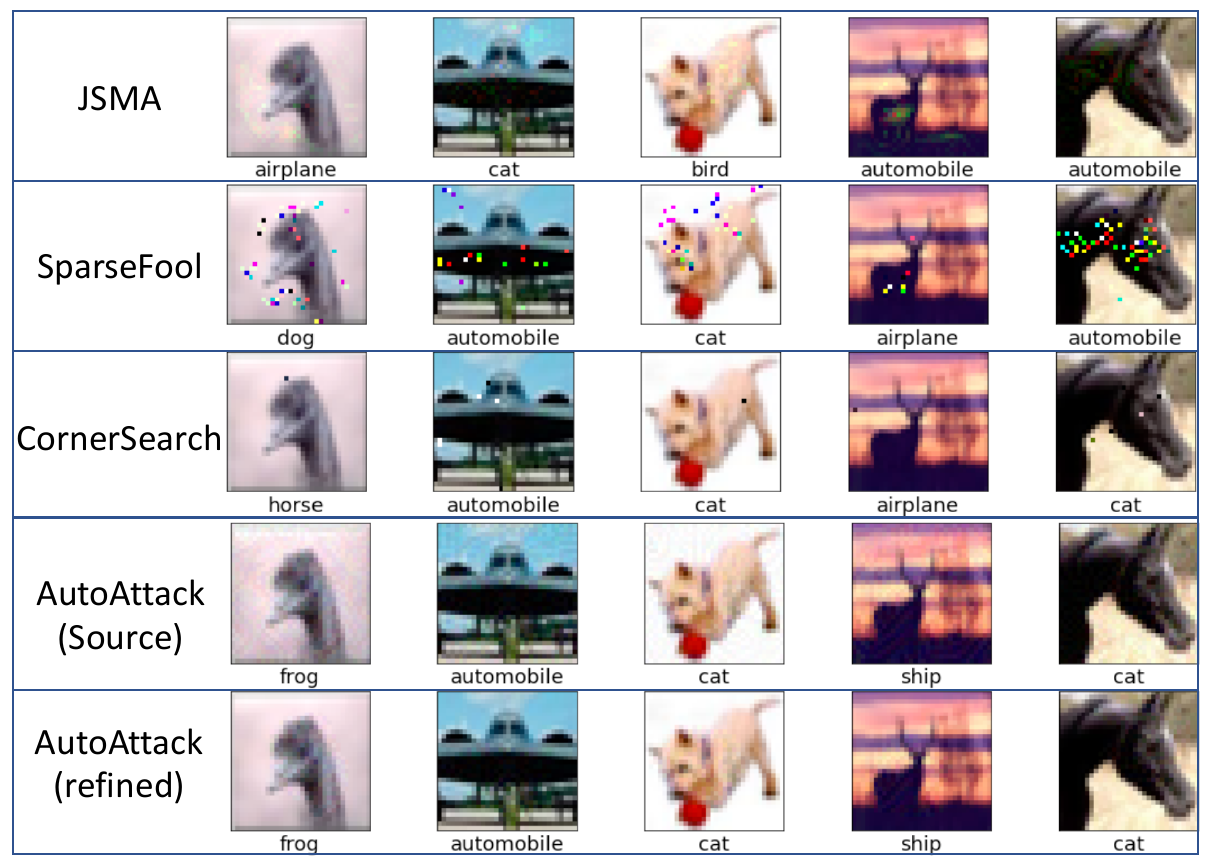}
         \end{subfigure} \\
             \begin{subfigure}[b]{0.99\linewidth}
                 \centering
                 \caption{}
        \resizebox{0.99\textwidth}{!}{
        \includegraphics{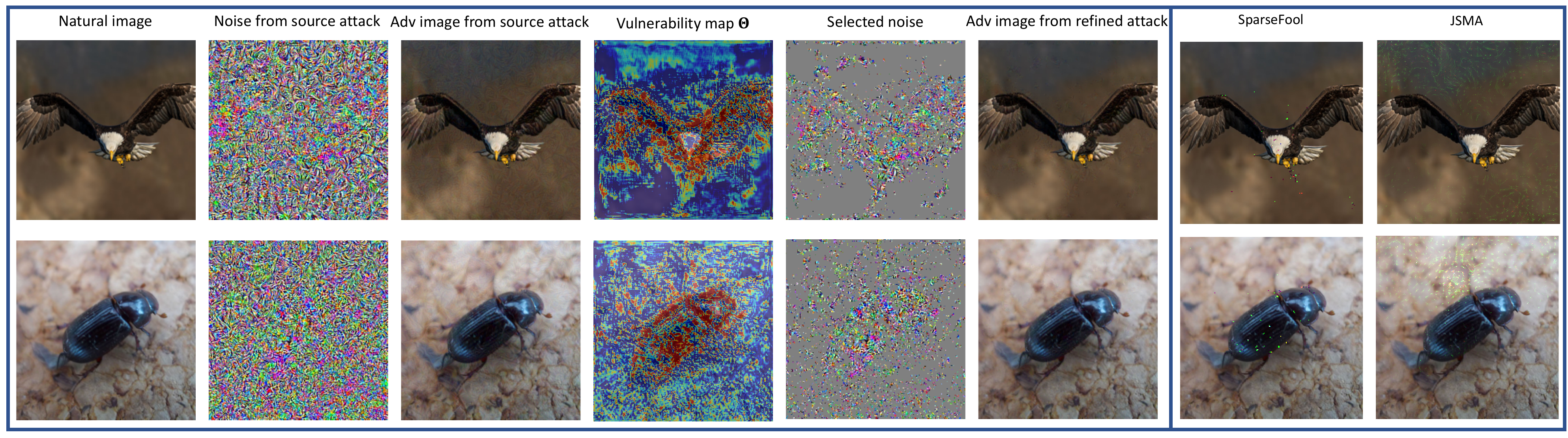}
        }
         \end{subfigure}
    \caption{(a) Visualisation of the source and refined attacks by PVAR on MNIST. (b) Sampled adversarial images on CIFAR10 with ResNet56 as the target classifier. All the attacks achieve similar AdvAcc as shown in Table~\ref{tb-sparse}. The labels below the images are the predicted ones of the classifier. (c) Left: Sampled images from ImageNet. We use PVAR with $\beta=0.3$ to refine BIM. Top image: original label ``bald eagle'', adv label ``prairie chicken''. Bottom image:  original label ``dung beetle'', adv label ``standard poodle''. Right: Adv images of JSMA and SparseFool. \textbf{Best viewed when zoomed-in.}}
    \label{fig-vis}
    \vspace{-3mm}
\end{figure*}

\textbf{Results of Comparison with Sparse Attacks.~}
Here we generate adv images from the test images of each dataset, with various sparse attacks and our refined dense attacks.
Unlike our approach, for some of the compared sparse attacks, controlling the exact number of perturbed pixels is infeasible. For fair comparison with our method, we gradually increase the regularisation parameters of the sparse attacks until their AdvAcc value on the training set is below 0.01\footnote{We are unable to further reduce AdvAcc for JSMA on ImageNet lower than 0.03 due to its efficiency.}. We use the same way of selecting $\beta=0.3$ for our PVAR. The detailed parameter settings of the compared sparse attacks are given in the supplementary material. The results are shown in Table~\ref{tb-sparse}. It can be observed that with similar attack performance, our refined attacks can achieve significantly lower DetAUC and $l_2$. Although sparse attacks can usually use fewer perturbations, the perturbation magnitudes can be large, resulting in larger DetAUC and $l_2$. On the other hand, ours is choosing from the perturbations from the source attacks, whose magnitudes are originally very small. We also report the averaged speed for attacking on one image. All the methods run on the same machine with an NVIDIA TITAN RTX GPU. PVAR clearly shows its faster attack speed, as it is several orders of magnitude faster than the optimisation-based sparse attacks. Even on high-resolution images like ImageNet, our approach can generate an adv image in less than 0.1 sec. Note that our attack time is the sum of those of the source attack and refinement processes, both of which can be done efficiently.

\begin{table}[!ht]
\vspace{-0.3cm}
\center
\caption{User study of perceivableness.}
\label{tb-us}
\resizebox{0.9\linewidth}{!}{
\begin{tabular}{@{}ccc@{}} 
 \toprule
                        & CIFAR10       & ImageNet      \\ \midrule
BIM                    & 76.66\%$\pm$5.56 & 82.68\%$\pm$3.66 \\ 
Our refined BIM        & 23.33\%$\pm$5.56 & 17.32\%$\pm$3.66 \\ \midrule
PGD                    & 77.32\%$\pm$3.66 & 81.34\%$\pm$5.60 \\ 
Our refined PGD        & 22.68\%$\pm$3.66 & 18.66\%$\pm$5.60 \\ \midrule
AutoAttack             & 78.68\%$\pm$8.70 & 77.34\%$\pm$7.60 \\ 
Our refined AutoAttack & 21.32\%$\pm$8.70 & 22.66\%$\pm$7.60 \\ \bottomrule
\end{tabular}
}
\vspace{-0.2cm}
\end{table}

\textbf{User Study.~}
For each dataset, we sampled 15 natural images and generated adversarial images with the source and our refined attacks. For each pair of two adversarial images, we asked each subject person to select \textit{the one that is with more perceivable perturbations}. We collected the responses from 15 people and then report the percentages of the choices averaged over the images for each attack, which are used as the score of ``perceivableness''.
\textit{A lower score indicates that an adversarial image is less perceivable to human, i.e., a better result}. The results are shown in Table~\ref{tb-us}. It can be seen that our method significantly reduces the ``the visual impact of the attacks.''

\textbf{Visualisation.~}
Figure~\ref{fig-vis}(a) shows the vulnerability map and source/refined attack of an MNIST digit 7. 
In Figure~\ref{fig-vis}(b), we qualitatively compare the adv images of our refined AutoAttack with other sparse attacks. One can observe that sparse attacks add sparse but large perturbations to natural images, which can be more easily detected by human eyes. For the original AutoAttack, although its noise is small, it spreads over the image. By removing ineffective perturbations of AutoAttack, our refined version looks significantly less noisy. Finally, we show the visualisation of ImageNet in Figure~\ref{fig-vis}(c), similarly to that in Figure~\ref{fig-demo}. It can be observed that the vulnerability maps intuitively help us understand the semantic structures of adversarial attacks.
Compared with the adv images of JSMA and SparseFool, the perturbations of our refined adv images are clearly less perceptible. More visualisations are provided in the supplementary material.

\section{CONCLUSION}
In this paper, we have tackled the task of attacking with fewer perturbations in a novel way, where we have proposed to refine given dense attacks by reducing their perturbations. 
Accordingly, we have proposed a probabilistic post-hoc framework that first uses a U-Net to learn the vulnerability map of a natural image and then selects from source attacks' perturbations based on the map. The framework is trained by the maximisation of the mutual information.
Our method can be applied to refine an arbitrary dense attack by removing 70\% of its perturbations in general. The refined attack keeps the same attack power as the original attack, but its adversarial images are significantly less detectable and closer to their natural counterparts.
Compared with sparse attacks, our refined attacks usually enjoy a smaller magnitude of perturbations, leading to better $l_2$ and DetAUC scores.
In addition, our method can generate adv images much more efficiently than sparse attacks.
We believe that the paper can inspire new thinking, and our study on vulnerability maps can assist the development of robust models by better protecting vulnerable pixels.

\bibliographystyle{apalike}
\bibliography{arxiv}

\appendix

\onecolumn

\section{LIST OF CONTENTS}

\begin{itemize}
    \item The derivation of Eq. (3) in the main paper.
    \item The architecture of U-Net.
    \item The settings of the sparse attacks.
    \item Study on the transferability of vulnerability maps.
    \item More visualisation on ImageNet. 
\end{itemize}

\section{DERIVATION OF EQ. (3) IN THE MAIN PAPER}

The derivation is shown in Eq.~(\ref{eq-mi-derive}).

\begin{align}
\label{eq-mi-derive}
&\mathbb{I}(\vec{y}^s, \matr{X}^{adv}) =
\int p(\vec{y}^s, \matr{X}^{adv}, \matr{X}^s, \matr{Z}, \matr{X}) \log \frac{p(\vec{y}^s, \matr{X}^{adv})}{p(\vec{y}^s) p(\matr{X}^{adv})} \text{d} \vec{y}^s \text{d} {\matr{X}^{adv}} \text{d} {\matr{X}^s} \text{d} \matr{Z} \text{d} \matr{X} \nonumber \\ 
&=\exptt{\matr{X}}{\exptt{\matr{X}^{s}| \matr{X}, \matr{Z} | \matr{X}}{\exptt{\matr{X}^{adv}|(\matr{X}^s, \matr{Z})}{\int p(\vec{y}^s|\matr{X}^s) \log \frac{p(\vec{y}^s|\matr{X}^{adv})}{p(\vec{y}^s)} \text{d}\vec{y}^s}}} \nonumber\\
&=\exptt{\matr{X}}{\exptt{\matr{X}^{s}|\matr{X}, \matr{Z}|\matr{X}}{\exptt{\matr{X}^{adv}|(\matr{X}^s,\matr{Z})}{\sum_{k=1}^{K} p(y_k|\matr{X}^s) \log p(y_k|\matr{X}^{adv})}}} + \text{constant},\\
\text{where we leverage }
&p(\vec{y}^s, \matr{X}^{adv}, \matr{X}^s, \matr{Z}, \matr{X}) = p(\vec{y}^s|\matr{X}^{s}) p(\matr{X}^{adv}|(\matr{X}^s, \matr{Z})) p(\matr{X}^s| \matr{X}) p(\matr{Z} | \matr{X}) p(\matr{X}). \nonumber
\end{align}

\section{ARCHITECTURE OF U-NET}
We apply a symmetric architecture for the U-Net~\cite{ronneberger2015u} used in our model to generate the vulnerability map $\matr{\Theta}$.
The U-Net consists of a down-sampling encoder and an up-sampling decoder. Specifically, the encoder is composed of 4 blocks each of which has:
    \begin{itemize}
        \item 3x3 convolution layer with batch normalization and ReLU activation
        \item 3x3 convolution layer with batch normalization and ReLU activation
       \item 3x3 convolution layer with batch normalization and ReLU activation
        \item 2x2 max-pooling
    \end{itemize}
The numbers of convolutional filters in the 4 blocks of the encoder are $[32, 64, 128, 256]$, respectively.

The decoder is similarly composed of 4 blocks, each of which has:
    \begin{itemize}
        \item 2x2 up-sampling layer using nearest neighbours
        \item Concatenation with the feature maps output by the corresponding block of the encoder
        \item 3x3 convolution layer with batch normalization and ReLU activation
        \item 3x3 convolution layer with batch normalization and ReLU activation
    \end{itemize}
The numbers of convolutional filters in the 4 blocks of the decoder are $[256, 128, 64, 32]$, respectively.

\section{SETTINGS OF THE SPARSE ATTACKS}
For the sparse attacks in our experiments, we use the default settings except the ones mentioned below:
\begin{itemize}
 \item JSMA~\cite{papernot2016limitations}: We set the attack strength $\epsilon$ to 0.3 for MNIST and 0.03 for CIFAR10/ImageNet, which is the same as the setting of the source attacks of PVAR. For the $\gamma$ parameter, which controls the number of pixels to attack, we gradually increase its value until JSMA's AdvACC is under 0.01. The specific setting of $\gamma$ on MNIST/CIFAR/ImageNet is 0.3/0.3/0.1 respectively.
 \item SparseFool~\cite{modas2019sparsefool}: We set the attack strength $\epsilon$ to 0.3 for MNIST and 0.03 for CIFAR10/ImageNet, which is the same as the setting of the source attacks of PVAR. For the $\lambda$ parameter, which controls the number of pixels to attack, we gradually increase its value until SparseFool's AdvACC is under 0.01. The specific settings of $\lambda$ on MNIST/CIFAR/ImageNet are 2.0/2.0/6.0 respectively.
 \item CornerSearch~\cite{croce2019sparse}: We set $K_{max}=\lceil 0.7 H W \rceil$, where $H$ and $W$ are the height and width of the images, respectively. We tune  $\kappa=1.5$ for MNIST and $\kappa=1.0$ for CIFAR10 to make its AdvACC value under 0.01. 
\end{itemize}
\section{TRANSFERABILITY OF VULNERABILITY MAPS}

In the experiments of the main paper, we use a source attack to train our model and use the trained model to refine the generate perturbations generated from the same source attack in the testing phase. To study the transferability of the vulnerability maps learned by our model across different attacks, we use a different attack to generate perturbations in the testing phase.
Table~\ref{tb-r32} shows the AdvAcc results on CIFAR10.
It can be observed that the vulnerability maps learned by our model from one dense attack can generalise to refine other attacks.

\begin{table}[]
\centering
\caption{Transferability  of vulnerability maps for ResNet32 on CIFAR10. AdvAcc is reported. ``Attacks for training'' means that the source attack that we use for training our model (i.e., the U-Net, $g_\alpha(\cdot)$) and ``Attacks for testing'' means that after $g_\alpha(\cdot)$ is trained, we use another attack to serve as the source attack to generate perturbations and then use the trained model to refine them in the testing phase. We set $\beta=0.3$.}
\label{tb-r32}
\begin{tabular}{|c|c|c|c|}
\hline
 \diagbox[width=10em]{Attacks \\for training}{Attacks \\for testing}   & PGD   & AT    & BIM   \\ \hline
PGD & 0.002 & 0.000 & 0.001 \\ \hline
AT  & 0.016 & 0.000 & 0.013 \\ \hline
BIM & 0.000 & 0.000 & 0.000 \\ \hline
\end{tabular}
\end{table}

\section{MORE VISUALISATIONS}
\begin{figure*}
    \centering
    \resizebox{0.999\textwidth}{!}{
    \includegraphics{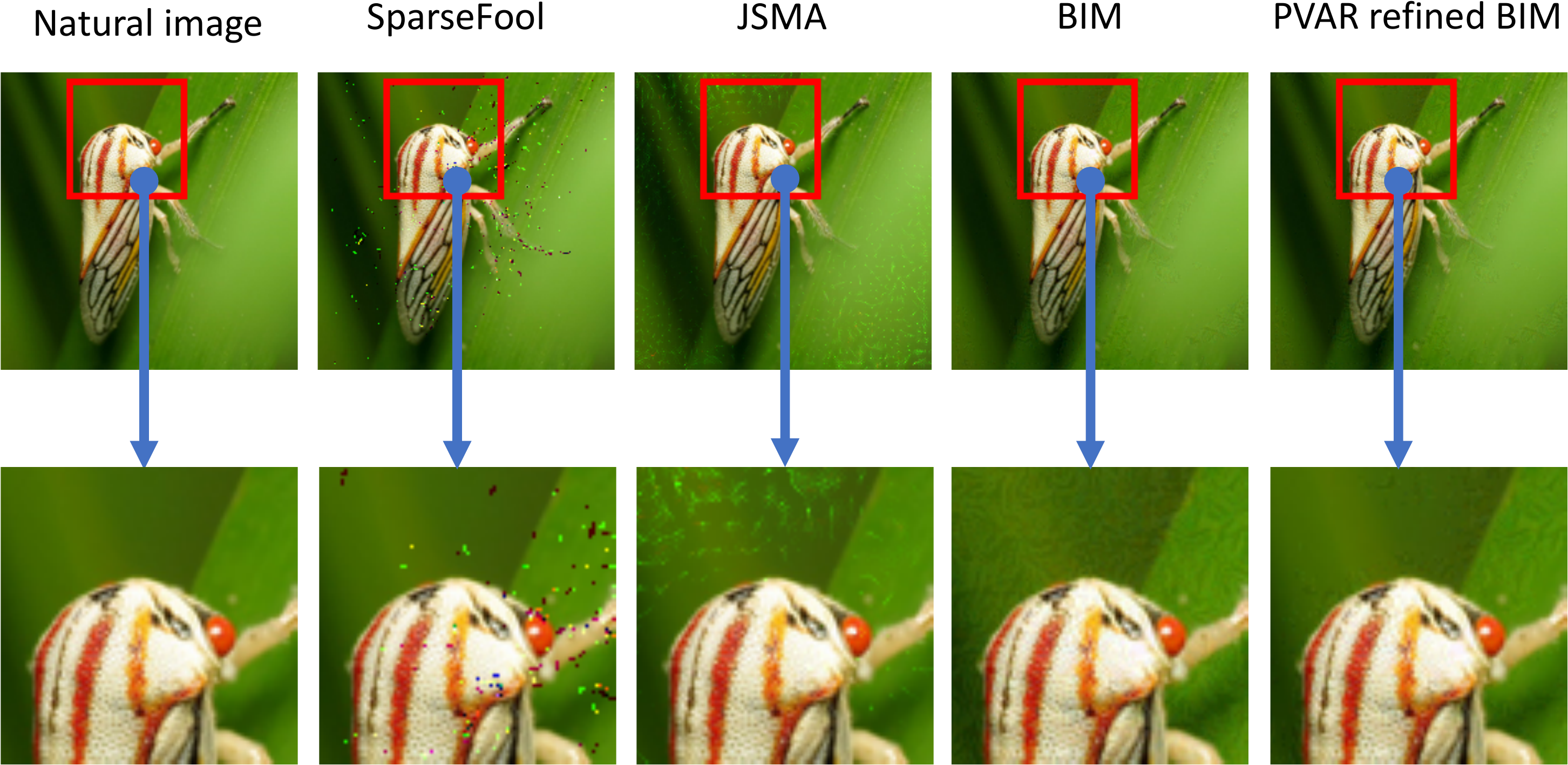}
    }
    \caption{}
    \label{fig-vis1}
\end{figure*}

\begin{figure*}
    \centering
    \resizebox{0.999\textwidth}{!}{
    \includegraphics{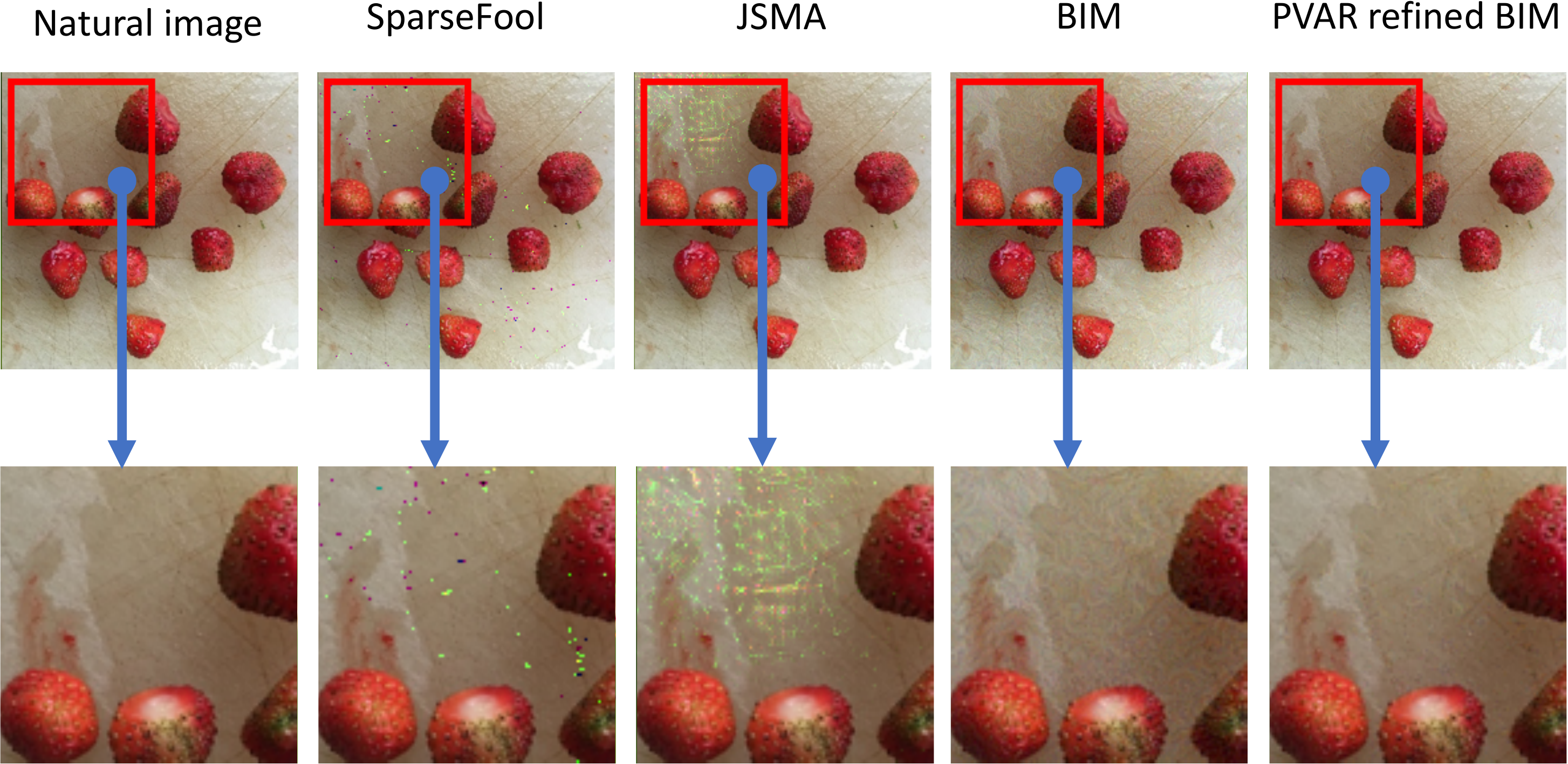}
    }
    \caption{}
    \label{fig-vis2}
\end{figure*}

\begin{figure*}
    \centering
    \resizebox{0.999\textwidth}{!}{
    \includegraphics{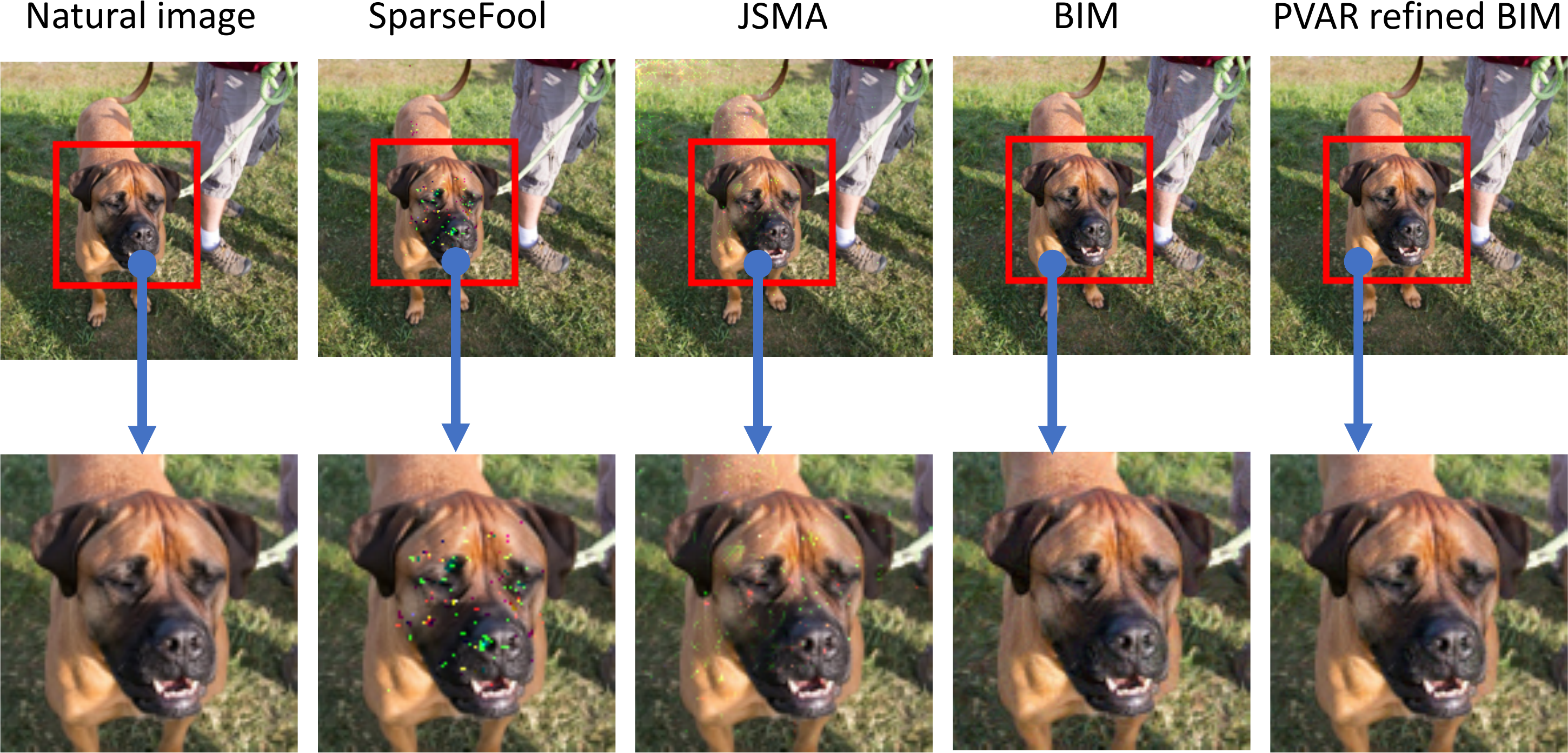}
    }
    \caption{}
    \label{fig-vis3}
\end{figure*}

In Figure~\ref{fig-vis1},~\ref{fig-vis2},~\ref{fig-vis3}, we show more ROI visualisations of the PVAR refined attacks and sparse attacks on ImageNet images.

\end{document}